\documentclass[11pt,letterpaper]{article}
\usepackage[nohyperref]{emnlp2017}
\usepackage{times}
\usepackage{latexsym}
\usepackage{url}

\usepackage{amsmath}
\usepackage{multirow}
\usepackage{siunitx} % Provides the \SI{}{} and \si{} command for typesetting SI units
\usepackage{graphicx} % Required for the inclusion of images
\usepackage{caption}
\usepackage{subcaption}
\usepackage{tikz}
\usepackage{pgfplots}
\usepackage{color,soul}
\usepackage{amssymb}
\usepackage{tablefootnote}
\usepackage{bm}
\usepackage{booktabs}
\usepackage{xspace}
\usepackage{footnote}
\usepackage{empheq}
\usepackage{algorithm}
\usepackage[noend]{algpseudocode}
\usepackage{adjustbox} % for rotating columns of tables
\usepackage{array}
\usepackage{etoolbox} % for siunitx with bold font
\newcommand{\ubold}{\fontseries{b}\selectfont} % renew def. for non-extended bold font
\robustify\ubold

\DeclareMathOperator*{\argmin}{arg\,min}

\newcommand{\vh}{\boldsymbol{h}}
\newcommand{\vx}{\boldsymbol{x}}
\newcommand{\vy}{\boldsymbol{y}}
\newcommand{\vr}{\boldsymbol{r}}

\newcommand{\vg}{\boldsymbol{g}}
\newcommand{\vW}{\boldsymbol{W}}
\newcommand{\vb}{\boldsymbol{b}}
\newcommand{\vc}{\boldsymbol{c}}
\newcommand{\valpha}{\boldsymbol{\alpha}}
\newcommand{\ve}{\boldsymbol{e}}

\newcommand{\vTheta}{\boldsymbol{\Theta}}

\newcommand{\norm}[1]{\left\lVert#1\right\rVert}
\newcommand\given[1][]{\:#1\vert\:}

\emnlpfinalcopy

\def\emnlppaperid{1250}

\title{ Towards Decoding as Continuous Optimisation in \\ Neural Machine Translation }

\author{Cong Duy Vu Hoang\\
	    University of Melbourne\\
	    Melbourne, VIC, Australia\\
	    {\tt vhoang2@student.unimelb.edu.au}
	  \AND
	Gholamreza Haffari\\
  	Monash University\\
  	Clayton, VIC, Australia\\
    {\tt gholamreza.haffari@monash.edu}
    \And
	Trevor Cohn\\
	University of Melbourne\\
	Melbourne, VIC, Australia\\
    {\tt t.cohn@unimelb.edu.au}
}

\date{}

\begin{document}

\maketitle

\begin{abstract}
We propose a novel decoding approach for neural machine translation (NMT) based on continuous optimisation. 
%transforms the decoding/inference of neural models into the online optimization problem. 
We convert  decoding -- basically a discrete optimization problem -- into a continuous optimization problem.  
The resulting constrained continuous optimisation problem is then tackled using gradient-based methods. % with Stochastic Gradient Descent (SGD) and Exponentiated Gradient (EG) methods.
%algorithms which are mainly used to train models in the literature. 
%
Our powerful decoding framework enables decoding intractable models such as the intersection of
left-to-right and right-to-left (bidirectional) as well as source-to-target and target-to-source (bilingual) NMT models.
Our empirical results show  that our decoding framework is  effective, 
and leads to substantial improvements in  translations generated from the intersected models where the typical greedy or beam search is not feasible. 
We also compare our framework against reranking, and analyse its advantages and disadvantages. 
 
%%% Local Variables: 
%%% mode: latex
%%% TeX-master: "acl2017-relopt.tex"
%%% End: 

\end{abstract}

\section{Introduction}
\label{sec::intro}

%brief intro about seq2seq background (training + decoding/inference)
Sequence to sequence learning with neural networks \cite{2013arXiv1308.0850G,Sutskever:2014:SSL:2969033.2969173,2015arXiv150600019L} is typically associated with two phases: training and decoding (\emph{a.k.a.} inference). 
Model parameters are learned by optimising the training objective, so that the model generalises well when the unknown test data is decoded. 
%
%The training phase is optimised in a way that during decoding, it generalises well over the unknown testing data.
The majority of literature have been focusing on developing better training paradigms or network architectures,  
but the decoding problem is arguably under-investigated. 
Conventional heuristic-based approaches for approximate inference include greedy, beam, and stochastic search.
%
%Conventionally, heuristics-based approaches are employed for approximate inference, including: greedy search, beam search, and stochastic search.  
%
Greedy and beam search have been empirically proved to be adequate for many sequence to sequence tasks, 
and are the standard methods for decoding in NMT. 

However, these approximate inference approaches have several drawbacks. 
Firstly, due to sequential decoding of symbols of the target sequence,  the inter-dependencies among the target symbols 
are not fully exploited. 
For example, when decoding the words of the target sentence in a left-to-right manner, 
the right context is not exploited leading potentially to inferior performance (see \newcite{Watanabe:2002:BDS:1072228.1072278} who apply this idea in traditional 
statistical MT).  %, leading to the difficulty of adding additional constraints for output symbols.  
%It has been shown for the traditional statistical MT models that unidirectional decoding is superior to 
 Secondly, it is not trivial to apply greedy or beam search to  decode in NMT models involving global features or constraints, e.g., intersecting left-to-right and right-to-left models which do not follow the same generation order.
These global constraints capture different aspects and can be highly useful in producing better and more diverse translations.

We introduce a novel decoding framework (\S~\ref{sec::prob_for}) that effectively relaxes this  \emph{discrete} 
optimisation problem  into a \emph{continuous} optimisation problem.
This is akin to linear programming relaxation approach for approximate inference in graphical models with discrete random variables 
where the exact inference is NP-hard \cite{Sontag_thesis10,Belanger:2016:SPE:3045390.3045495}. 
Our continuous optimisation problems are challenging due to the non-linearity and non-convexity 
of the relaxed decoding objective.
We make use of  stochastic gradient descent (SGD) and exponentiated gradient (EG) algorithms, which 
are mainly used for training  in the literature, for  decoding based on our relaxation approach.
 Our decoding framework  is powerful and flexible, as it enables us to decode with global constraints involving intersection of multiple NMT 
 models (\S \ref{sec::constraints}).
 We present experimental results on Chinese-English and German-English translation tasks, confirming the effectiveness of our relaxed optimisation method for decoding
% framework as well as the empirical analysis of the aforementioned optimisation algorithm for decoding 
(\S \ref{sec::exp}).%
%\footnote{The source code can be accessed in  upon publication.} 
 %leading to substantial improvement in BLEU score.
 %
  
%However, our decoding in neural models is more challenging than that in graphical models since the resulting 
%optimisation problems are usually non-linear and non-convex, due to the usage of  non-linear 
%activation functions in deep neural architectures. 
%
%We relax the integrality constraints of the inference problem to formulate it as a continuous optimisation problem, which
%s then optimised using stochastic gradient descent and its variants. 
%
%These optimisation algorithms have been mainly used for training of neural architectures in the literature, whereas
%we utilise them for approximate inference. 

%Our approximate inference framework  is quite powerful and flexible, as it allows to decode for an output which 
%jointly optimises the scores of several models, 
%e.g., decoding by intersecting left-to-right and right-to-left neural translation models, 
%intersecting source-to-target and target-to-source models in a principled way. 
%
%Our experimental results (Section~\ref{sec::exp}) show the promise of the proposed framework.

%%% Local Variables: 
%%% mode: latex
%%% TeX-master: "acl2017-relopt"
%%% End: 

\section{Neural Machine Translation} 
\label{sec::am}
We briefly review the attentional neural translation model proposed by \newcite{bahdanau:ICLR:2015} as a sequence-to-sequence neural model onto which we will apply our decoding framework. 

In neural machine translation (NMT), the probability of the target sentence $\vy$ given a source sentence $\vx$ is written as:
\begin{eqnarray} \label{eqn::nmt_1}
\operatorname{P}_{\vTheta} \left( \vy | \vx \right) &=& \sum_{i=1}^{|\vy|} \log \operatorname{P}_{\vTheta} \left( y_i | \vy_{<i}, \vx \right) \\
 y_i | \vy_{<i}, \vx  &\sim&  \operatorname{softmax} \left( \pmb{f}( \vTheta, \vy_{<i}, \vx ) \right) \nonumber
\end{eqnarray}
where $\operatorname{f}$ is a non-linear function of the previously generated sequence of words $\vy_{<i}$, the source sentence $\vx$, and the model parameters $\vTheta$.
In this paper, we realise $\pmb{f}$ as follows:
\begin{equation}
\begin{aligned}
\pmb{f}( \vTheta, \vy_{<i}, \vx) &= \vW_{o} \cdot \operatorname{MLP} \left( \vc_i, \pmb{E}_T^{y_{i-1}}, \vg_i \right) + \vb_o \\
\vg_i &= \operatorname{RNN}^{\phi}_{dec} \left( \vc_i, \pmb{E}_T^{y_{i-1}}, \vg_{i-1} \right) \nonumber
\end{aligned}
\end{equation}
where $\operatorname{MLP}$ is a single hidden layer neural network with $\operatorname{tanh}$ activation function, and 
$\pmb{E}_T^{y_{i-1}}$ is the embedding of the target word $y_{i-1}$ in the embedding matrix $\pmb{E}_T \in \mathbb{R}^{n_e \times |V_T|}$ of the target language vocabulary $V_T$ and $n_e$ is the embedding dimension.
The state  $\vg_i$ of the decoder RNN is a function of $y_{i-1}$, its previous state $\vg_{i-1}$,
and the \emph{context} $\vc_i=\sum_{j=1}^{|\vx|} \alpha_{ij} \vh_j $ summarises parts of the source sentence which are \emph{attended} to, where
\begin{equation*}\label{eqn::nmt_2}
\begin{aligned}
%\vc_i &= \sum_{j=1}^{|\vx|} \alpha_{ij} \vh_j \nonumber \\
\valpha_i &= \operatorname{softmax}(\ve_i) \quad \textrm{;} \quad 
e_{ij} = \operatorname{MLP} \left( \vg_{i-1}, \vh_j \right) \\
\vh_j &= \operatorname{biRNN}^{\theta}_{enc} \left( \pmb{E}_S^{x_j},   \overrightarrow{\vh}_{j-1}, \overleftarrow{\vh}_{j+1}  \right)
\end{aligned}
\end{equation*}
In above, $\overrightarrow{\vh}_i$  and $\overleftarrow{\vh}_i$ are the states of the left-to-right and right-to-left RNNs encoding the source sentence, 
and $\pmb{E}_S^{x_j}$ is the embedding of the source word $x_j$ in the embedding matrix $\pmb{E}_S \in \mathbb{R}^{n'_e \times |V_S|}$ of the source language vocabulary $V_S$ and $n'_e$ is the embedding dimension.

Given a bilingual corpus $\mathcal{D}$, the model parameters are learned by maximizing the (regularised) conditional log-likelihood:
\begin{equation}\label{eqn::nmt_3}
\vTheta^* := \operatorname{argmax}_{\vTheta}  \sum_{(\vx,\vy) \in \mathcal{D}} \log \operatorname{P}_{\vTheta} \left( \vy \given \vx \right).
\end{equation}
%usually using stochastic gradient descent (SGD) or its variants. %In this paper, we are interested in the decoding problem though which is outlined in the next section. 
%
The model parameters $\vTheta$ include the weight matrix
$\vW_{o} \in \mathbb{R}^{|V_T| \times n_h}$ and the bias $\vb_o \in \mathbb{R}^{|V_T|}$
-- with $n_H$ denoting the hidden dimension size --
as well as the RNN encoder $\operatorname{biRNN}^{\theta}_{enc}$ / decoder $\operatorname{RNN}^{\phi}_{dec}$ parameters, word embedding matrices, and those of the attention mechanism.
The model is trained end-to-end by optimising the training objective using stochastic gradient descent (SGD) or its variants. 
In this paper, we are interested in the decoding problem though which is outlined in the next section.

%Here, it's worth noting that the function $\operatorname{f}$ in Equation~\ref{eqn-exp-am-train-2} will be used directly in our proposed RelOpt framework as mentioned in Equation~\ref{eqn-fm-4a}.

%\textbf{Decoding}. For decoding, simple heuristics-based methods (greedy Viterbi or beam search) can be employed; and the decoding results (translations) can be obtained sequentially (e.g., left-to-right).

%\begin{equation}\label{eqn-exp-am-train-1}
%\begin{aligned}
%\vTheta &:= \operatorname{argmax}_{\vTheta} \{ \operatorname{P}_{\vTheta} \left( \vy \given \vx \right) \} \nonumber \\
%&:= \operatorname{argmax}_{\vTheta} \{ \log \operatorname{P}_{\vTheta} \left( \vy \given \vx \right) \} \\
%&:= \operatorname{argmax}_{\vTheta} \{ \sum_{i} \log \operatorname{P}_{\vTheta} \left( y_i | \vx \right) \};
%\end{aligned}
%\end{equation}
%where the conditional probability $\operatorname{P}_{\vTheta} \left( y_i | \vx \right)$ of generating the target word $y_i$ given the source sequence $\vx$ is formulated as:
%\begin{equation}\label{eqn-exp-am-train-2}

%{\small
\begin{algorithm*}[t]
\begin{algorithmic}[1]
 %\SetAlgoLined
 %\KwData{ an input sequence $\vx$; pre-trained model parameters $\vTheta$; pre-defined target sequence length $\ell$; target vocabulary $V_T$; pre-defined iteration number $N$. }
 \State For all $i$ initialise $\hat{\vy}^0_i \in \Delta_{|V_T|}$
 %\State{Choose a learning rate $\eta > 0$}
 \For{ $t = 1, \ldots, \textrm{MaxIter}$} \Comment{ $Q(.)$ is defined as eqn (\ref{eqn:q_def})}
 \State{For all $i, w :  \textrm{calculate } \nabla_{i,w}^{t-1} = \frac{\partial Q(\hat{\vy}_1^{t-1},\ldots,\hat{\vy}^{t-1}_{\ell})}{\partial \hat{\vy}_i(w)}$} \Comment{ using backpropagation}
 \State{For all $i, w :  \textrm{update } \hat{\vy}_{i}^t(w) \propto  \hat{\vy}_{i}^{t-1}(w) \cdot \exp\left(- \eta \nabla_{i,w}^{t-1} \right)$ } \Comment{$\eta$ is the step size} 
  \EndFor
 \State  \Return $\arg\min_t Q(\hat{\vy}_1^t,\ldots,\hat{\vy}_{\ell}^t)$ %{$\{\hat{\vy}^{*}\}_{i=1}^{\ell}$ with minimum cost ${Q}$. }
 \end{algorithmic}
\caption{The EG Algorithm for Decoding by Optimisation}
 \label{alg1a}
\end{algorithm*}

\section{Decoding as Continuous Optimisation}
\label{sec::prob_for}
%decoding/inference
In decoding, we are interested in finding the highest probability translation for a given source sentence: 
\begin{equation}\label{eqn-fm-1}
\begin{aligned} 
\text{minimise}_{\vy} \quad  -P_{\vTheta} \left( \vy \given \vx \right) \quad \text{s.t.} \quad  \vy \in \mathcal{Y}_{\vx}
\end{aligned}
\end{equation}
where $\mathcal{Y}_{\vx}$ is the space of possible translations for the source sentence $\vx$.
In general, searching $\mathcal{Y}_{\vx}$ to find the highest probability translation is intractable due to long-range dependency terms in eqn (\ref{eqn::nmt_1}) which prevents dynamic programming for efficient search algorithms in this exponentially-large space of possible translations with respect to the input length $|\vx|$.

We now formulate this discrete optimisation problem as a continuous one, and then use standard algorithms for continuous optimisation for decoding.  
Let us assume that the maximum length of a possible translation for a source sentence is known and denote it by $\ell$.  
The best translation for a given source sentence solves the following optimisation problem: 
\begin{align}
\vy^{*} 
%&= \argmin_{\vy} -\log \operatorname{P}_{\vTheta} \left( \vy \given \vx \right)  \\
%&= \argmin_{y_1,\ldots, y_{\ell}} \prod_{i=1}^{\ell} -\operatorname{P}_{\vTheta} \left( y_{i} \given \vy_{<i}, \vx  \right) \\
= &\argmin_{y_1,\ldots, y_{\ell}} \sum_{i=1}^{\ell}  -\log\operatorname{P}_{\vTheta} \left( y_{i} \given \vy_{<i}, \vx  \right) \label{eqn-fm-2} \\
&\text{s.t.}  \quad  \forall i \in \{1 \ldots \ell \} : y_i \in V_T. \nonumber
\end{align}
%where $V_T$ is  the target language vocabulary. 
Equivalently, we can re-write the above discrete optimisation problem as follows:
\begin{align}
%\begin{aligned}
%\vy^{*} %&= \argmin_{\tilde{\vy}_1,\ldots, \tilde{\vy}_{\ell}} - \sum_{i=1}^{\ell}  \log\operatorname{P}_{\vTheta} \left( \tilde{\vy}_{i} \given \tilde{\vy}_{<i}, \vx  \right) \\
  &\argmin_{\tilde{\vy}_1,\ldots, \tilde{\vy}_{\ell}} - \sum_{i=1}^{\ell} \tilde{\vy}_i  \cdot \log  \operatorname{softmax} \left( \pmb{f} \left( \vTheta, \tilde{\vy}_{<i}, \vx \right) \right)  \nonumber\\
&\text{s.t.}  \quad  \forall i \in \{1 \ldots \ell \} : \tilde{\vy}_i \in \mathbb{I}^{|V_T|}  %\nonumber
\label{eqn-fm-3}
\end{align}
where $\tilde{\vy}_i$ are vectors using the one-hot representation of the target words $\mathbb{I}^{|V_T|} $.  
%
%As part of this change of variables from $y_i$ to $\tilde{\vy}_i$, the embedding term $\pmb{E}_T^{y_i}$ is replaced by its equivalent $\pmb{E}_T \cdot \tilde{\vy}_i$ in the neural network computation.
% Trevor -- commented out above

We now convert the optimisation problem (\ref{eqn-fm-3}) to a continuous one by dropping the integrality constraints $\tilde{\vy}_i \in \mathbb{I}^{|V|} $ and require the variables to take values from the probability simplex:
\[ \boxed{ \begin{aligned}
%\hat{\vy}^{*} 
 \argmin_{\hat{\vy}_1,\ldots, \hat{\vy}_{\ell}} & - \sum_{i=1}^{\ell} \hat{\vy}_i  \cdot { \log  \operatorname{softmax}} \left( \pmb{f} \left( \vTheta, \hat{\vy}_{<i}, \vx \right) \right)
\\
& \text{s.t.}  \quad  \forall i \in \{1 \ldots \ell \} : \hat{\vy}_i
\in \Delta_{|V_T|}
\end{aligned} } \]
where $\Delta_{|V_T|}$ is the $|V_T|$-dimensional probability simplex, i.e., \mbox{$\{ \hat{\vy}_i \in [0,1]^{|V_T|} :  \norm{\hat{\vy}_i}_1 = 1\}$.} 
Intuitively, this amounts to replacing  $\pmb{E}_T^{y_i}$ with  the \emph{expected} embedding of  target language words  $\mathbb{E}_{\hat{\vy}_i(w)}[\pmb{E}_T^{w}]$ under the distribution $\hat{\vy}_i$ in the NMT model.
%= \sum_{w \in V_T} \hat{\vy}_i(w) \pmb{E}_T^{w}$ to replace $\pmb{E}_T^{y_i}$ with 

After  solving the above constrained continuous optimisation problem, there is no guarantee that the resulting solution $\{\hat{\vy}_i^{*}\}_{i=1}^{\ell}$ to include one-hot vectors corresponding to target language words. It instead will have \emph{distributions} over target language vocabulary for each random variable of interest in  prediction, so we need a technique to round up this \emph{fractional} solution. 
Our method is to put all of the probability mass on the word with the highest probability\footnote{If there are multiple words with the same highest probability mass, we choose one of them arbitrarily.} for each $\hat{\vy}_i^*$. % and denote the resulting discrete solution by $\tilde{\vy}^{*}$. 
We leave  exploration of more elaborate projection techniques to the future work. 

In the context of graphical models, the above relaxation technique gives rise to linear programming for approximate inference \cite{Sontag_thesis10,Belanger:2016:SPE:3045390.3045495}. 
However, our decoding problem is much harder due to the non-linearity and non-convexity of the objective function operating on high dimensional space for deep models. 
We now turn our attention to optimisation algorithms to effectively solve the decoding optimisation problem. % (\ref{eqn-fm-4}). 

\subsection{Exponentiated Gradient (EG)}
Exponentiated gradient \cite{Kivinen:1997:EGV:253138.253139} is an elegant algorithm for solving optimisation problems involving simplex constraints. 
Recall our constrained optimisation problem: 
\begin{align*}
& \arg\min_{\hat{\vy}_1,\ldots,\hat{\vy}_{\ell}} Q(\hat{\vy}_1,\ldots,\hat{\vy}_{\ell}) \\
& \text{s.t.}  \quad \forall i \in \{1 \ldots \ell \} : \hat{\vy}_i \in \Delta_{|V_T|}
\end{align*}
where $Q(\hat{\vy}_1,\ldots,\hat{\vy}_{\ell})$ is defined as 
\begin{align} 
 - \sum_{i=1}^{\ell} \hat{\vy}_i  \cdot { \log  \operatorname{softmax}} \left( \pmb{f} \left( \vTheta, \hat{\vy}_{<i}, \vx \right) \right).
\label{eqn:q_def}
\end{align}
EG is an iterative algorithm, which updates each distribution $\hat{\vy}_i^t$ in the current time-step $t$ based on the distributions of the previous time-step as follows:
\begin{align*}
\forall w \in V_T: \quad \hat{\vy}_{i}^t(w) =\frac{1}{Z_i^t}   \hat{\vy}_{i}^{t-1}(w) \exp\left(- \eta \nabla_{i,w}^{t-1} \right)
\end{align*}
where $\eta$ is the step size, $\nabla_{i,w}^{t-1} = \frac{\partial Q(\hat{\vy}_1^{t-1},\ldots,\hat{\vy}^{t-1}_{\ell})}{\partial \hat{\vy}_i(w)}$ and $Z_i^t$ is the normalisation constant
$$Z_i^t = \sum_{w \in V_T}  \hat{\vy}_{i}^{t-1}(w) \exp\left(- \eta \nabla_{i,w}^{t-1} \right).$$
The partial derivatives $\nabla_{i,w}$ are calculated using the back propagation algorithm treating $\hat{\vy}_i$'s as \emph{parameters} and the
original parameters of the model $\vTheta$ as constants.
Adapting EG to our decoding problem leads to Algorithm \ref{alg1a}. 
It can be shown that the EG algorithm is a gradient descent algorithm for minimising the following objective function subject to the simplex constraints:
\begin{align}
& Q(\hat{\vy}_1,\ldots,\hat{\vy}_{\ell}) - \gamma \sum_{i=1}^{\ell}  \sum_{w \in V_T} \hat{\vy}_i(w)  \log \frac{1}{\hat{\vy}_i(w)} \nonumber \\
& = Q(\hat{\vy}_1,\ldots,\hat{\vy}_{\ell}) - \gamma \sum_{i=1}^{\ell} \textrm{Entropy}(\hat{\vy}_i) \label{eqn:eq_obj}
\end{align}
% the objective function endowed with the negative entropy: 
%\begin{align} 
%\sum_{i=1}^{\ell} & - \hat{\vy}_i \cdot \log \big[ \operatorname{softmax} \left( \operatorname{f} \left(\vTheta, \hat{\vy}_{<i}, \vx \right) \right) \big] 
 %\label{eqn-alg-1}  \\
%& - \gamma \hat{\vy}^T_i \log \frac{1}{\hat{\vy}_i} \quad \text{s.t.}  \quad  \forall i \in \{1 \ldots \ell \} : \vy_i \in \Delta_{|V|}.     \nonumber
%\end{align}
In other words, the algorithm looks for the maximum entropy solution which also maximizes the log likelihood under the model.  
There are intriguing parallels with the maximum entropy formulation of log-linear models \cite{Berger:1996:MEA:234285.234289}. %as well as minimising entropy in semi-supervised learning \cite{Grandvalet:2004:SLE:2976040.2976107}. 
In our setting, the entropy term acts as a prior which discourages overly-confident estimates without sufficient evidence. 

%{\small
\begin{algorithm*}[t]
\begin{algorithmic}[1]
 %\SetAlgoLined
 %\KwData{ an input sequence $\vx$; pre-trained model parameters $\vTheta$; pre-defined target sequence length $\ell$; target vocabulary $V_T$; pre-defined iteration number $N$. }
 \State For all $i$ initialise $\hat{\vr}^0_i$
 %\State{Choose a learning rate $\eta > 0$}
 \For{ $t = 1, \ldots, \textrm{MaxIter}$} \Comment{$Q(.)$ is defined in eqn (\ref{eqn:q_def}) 
and $\hat{\vy}_i = \operatorname{softmax}(\hat{\vr}_i)$ }
 \State{For all $i, w :  \textrm{calculate } \nabla_{i,w}^{t-1} = \sum_{w' \in V_T} \frac{\partial Q(\hat{\vy}_1^{t-1},\ldots,\hat{\vy}^{t-1}_{\ell})}{\partial \hat{\vy}_i(w')} \frac{\partial \hat{\vy}_i(w')}{\partial \hat{\vr}_i(w)}$} \Comment{ using backpropagation}
 \State{For all $i, w :  \textrm{update } \hat{\vr}_{i}^t(w) =  \hat{\vr}_{i}^{t-1}(w) - \eta \nabla^{t-1}_{i,w}$ } \Comment{$\eta$ is the step size} 
  \EndFor
 \State  \Return $\arg\min_t Q(\operatorname{softmax}(\hat{\vr}_1^t),\ldots,\operatorname{softmax}(\hat{\vr}_{\ell}^t))$ %{$\{\hat{\vy}^{*}\}_{i=1}^{\ell}$ with minimum cost ${Q}$. }
 \end{algorithmic}
\caption{The SGD Algorithm for Decoding by Optimisation}
 \label{alg2a}
\end{algorithm*}

%\paragraph*{The \textit{Softmax} Algorithm.} 
\subsection{Stochastic Gradient Descent (SGD) }
To be able to apply SGD to our optimisation problem, we need to make sure that the simplex constraints are kept intact. 
One way to achieve this  is by changing the optimisation variables from $\hat{\vy}_i$ to $\hat{\vr}_i$ through the $\operatorname{softmax}$ transformation, i.e. $\hat{\vy}_i = \operatorname{softmax} \left( \hat{\vr}_i \right)$. 
The resulting \emph{unconstrained} optimisation problem then becomes
\begin{equation*}
{\small
\begin{aligned}
\argmin_{\hat{\vr}_1,\ldots, \hat{\vr}_{\ell}} - \sum_{i=1}^{\ell} \operatorname{softmax} \left( \hat{\vr}_i \right)  \cdot \log  \operatorname{softmax} \left( \operatorname{f} \left( \vTheta, \hat{\vy}_{<i}, \vx \right) \right)
\end{aligned}
}
\end{equation*}
where $\pmb{E}_{T}^{y_i}$ is replaced with the expected embedding of the target words under the distribution resulted from the $\mathbb{E}_{\operatorname{softmax}(\hat{\vr}_i)}\left[\pmb{E}_T^w\right]$ in the model.

To apply SGD updates, we need the gradient of the objective function with respect to the new variables $\hat{\vr}_i$ which can be derived with the back-propagation algorithm based on the chain rule: 
\begin{equation*}
 \frac{\partial {Q}}{\partial \hat{\vr}_i(w)} = \sum_{w' \in V_T} \frac{\partial Q(.)}{\partial \hat{\vy}_i(w')} \frac{\partial \hat{\vy}_i(w')}{\partial \hat{\vr}_i(w)} %= \nabla_{\hat{\vy}} \mathbf{Q} \left( \hat{\vy} \right) \nabla_{\hat{\vr}} \left( \hat{\vy} \right). 
\end{equation*}
The resulting SGD algorithm is summarized in Algorithm \ref{alg2a}.

\section{Decoding in Extended NMT}
\label{sec::constraints}

Our decoding framework allows us to effectively and flexibly add additional global factors over the output symbols during inference. 
This in enabling by allowing decoding for richer global models, for which there is no effective means of  
greedy decoding or beam search.
We outline several such models, and their corresponding relaxed objective functions  for optimisation-based decoding.

\paragraph{Bidirectional Ensemble.}

Standard NMT generates the translation in a left-to-right manner, conditioning each target word on its left context. 
However, the joint probability of the translation can be decomposed in a myriad of different orders; one compelling alternative would be to condition each target word on its right context, i.e., generating the target sentence from right-to-left. 
We would not expect a right-to-left model to outperform a left-to-right, however, as the left-to-right ordering reflects the natural temporal order of spoken language.
However, the right-to-left model is likely to provide a complementary signal in translation as it will be bringing different 
biases and making largely independent prediction errors to those of the left-to-right model.
For this reason, we propose to use both models, and seek to find translations that have high probability according both models (this mirrors work on bidirectional decoding in classical statistical machine translation by \newcite{watanabe2002bidirectional}.)
Decoding under the ensemble of these models leads to an intractable search problem, not well suited to traditional greedy or beam search algorithms, which require a fixed generation order of the target words. 
This ensemble decoding problem can be formulated simply in our linear relaxation approach, using the following objective function:
\begin{align} 	
\mathcal{C}_{+\text{bidir}} := 
& -\alpha \log\operatorname{P}_{\vTheta_{\leftarrow}} \left({\vy} \given \vx \right) \nonumber \\
& - (1 - \alpha) \log\operatorname{P}_{\vTheta_{\rightarrow}} \left({\vy} \given \vx \right);
\label{eqn-sc-3}
\end{align}
where $\alpha$ is an interpolation hyper-parameter, which we set to 0.5; $\vTheta_{\rightarrow}$ and 
$\vTheta_{\leftarrow}$ are the pre-trained left-to-right and right-to-left models, respectively. 
This bidirectional agreement may also lead to improvement in translation diversity, as shown in \cite{2016arXiv160100372L} in a re-ranking evaluation.

\paragraph{Bilingual Ensemble.}
Another source of complementary information is in terms of the translation direction, 
that is forward translation from the source to the target language, and reverse translation in the target to source direction.
The desire now is to find a translation which is good under both the forward and reverse translation models.
This is inspired by the direct and reverse feature functions commonly used in classical discriminative SMT \cite{och2002discriminative} which have been shown to offer some complementary benefits (although see \cite{lopez2006word}).
%
%Decoding to find a translation which maximises the combination of the translation which is good under both forward and reverse translation models 
%is clearly an intractable search problem, as the reverse model is global over the target language, meaning there is no obvious means of search.
%
More specifically, we decode for the best translation in the intersection of the source-to-target and target-to-source models
by minimizing the following objective function:
%
%the relaxed objective function for optimisation-based decoding can be 
%expressed for the intersection of source-to-target and target-to-source models, formulated as:
\begin{align}
\mathcal{C}_{+\text{biling}} := & -\alpha \log\operatorname{P}_{\vTheta_{\textrm{s}\rightarrow\textrm{t}}} \left({\vy} \given \vx \right) \nonumber \\
& - (1 - \alpha) \log\operatorname{P}_{\vTheta_{\textrm{s}\leftarrow\textrm{t}}} \left({\vx} \given \vy \right); %\\
%&-\gamma \operatorname{tr}\left(\valpha_{\textrm{s}\rightarrow\textrm{t}} \valpha^{T}_{\textrm{s}\leftarrow\textrm{t}}\right);
\label{eqn-sc-4}
\end{align}
% \begin{equation}\label{eqn-sc-4a}
% \begin{aligned} 	
% &\mathcal{C}_{+biling} := \\
% &- \alpha \frac{1}{|\vy|} \log\operatorname{P}_{\vTheta_{st}} \left({\vy} \given \vx \right) \\
% &- (1 - \alpha) \frac{1}{|\vx|} \log\operatorname{P}_{\vTheta_{ts}} \left({\vx} \given \vy \right) \\
% &- \gamma \operatorname{trace\_bonus} \left( \valpha_{st}, \valpha^{T}_{ts} \right);
% \end{aligned}
% \end{equation}
where $\alpha$ is an interpolation hyper-parameter to be fine-tuned; and $\vTheta_{\textrm{s}\rightarrow\textrm{t}}$ and $\vTheta_{\textrm{s}\leftarrow\textrm{t}}$ are the pre-trained source-to-target and target-to-source models, respectively.
%Inspired by \newcite{cohn-EtAl:2016:N16-1}, the last term in (\ref{eqn-sc-4}) is the  so-called trace bonus which encourages agreement between the two models' alignments. 
%
Decoding for the best translation under the above objective function leads to an intractable search problem, 
as the reverse model is global over the target language,
meaning there is no obvious means of search with greedy algorithm or alike. 

\paragraph{Discussion.}
%Our framework is highly general, and several other model architectures can be easily supported, such as the noisy-channel model \cite{Koehn:2010:SMT:1734086}, which combines source to target translation with a target language model; 
%or decoding with additional constraints, e.g., source coverage \cite{DBLP:conf/icml/XuBKCCSZB15,mi-EtAl:2016:EMNLP2016} or word fertility \cite{cohn-EtAl:2016:N16-1}.\footnote{These constraints are only enforced only during training but not decoding, presumably due to intractable search. Our framework could allow their use in decoding.}
%
%We leave these and other extensions for future work.

There are two important considerations on how best to initialise the relaxed optimisation in the above settings, 
and how best to choose the step size.
As the relaxed optimisation problem is, in general, non-convex, finding a plausible initialisation is likely to be important for avoiding local optima.
Furthermore, a proper step size  is a key in the success of the EG-based and SGD-based  optimisation algorithms, 
and there is no obvious method how to best choose its value. 
We may also adaptively change the step size using (scheduled) annealing or via  the line search. 
We return to this considerations in the experimental evaluation.

% \subsection{Joint Decoding with a Language Model}
% We can intersect a language model with a neural translation model, in the spirit of the noisy channel model for traditional SMT \cite{Koehn:2010:SMT:1734086}:
% \begin{equation} \label{eqn-sc-5}
% \begin{aligned}	
% &\mathcal{C}_{+monoLM} := \\
% &-\log\operatorname{P}_{\vTheta} \left( \vy \given \vx  \right) - \lambda \log\operatorname{P}_\textrm{LM} \left( \vy \right);
% \end{aligned}
% \end{equation}
% where $\lambda$ is a hyper-parameter, weighting the importance of the additional monolingual language model, and 
% %
% $\operatorname{P}_\textrm{LM} \left( \vy \given \vx  \right)$ is the target language model (e.g., N-gram \cite{Koehn:2010:SMT:1734086}, NPLM \cite{Bengio:2003:NPL:944919.944966}, RNNLM \cite{MnihHinton2007,milokov-phdthesis,milokov:ASRU}). 

% It's worth noting that the additional monolingual language model can play a flexible role as a domain adapter for a neural MT system. 
% For example, the neural MT system can be trained on the news domain, whereas the additional monolingual target language model can be trained on another domain of interest, e.g., the conversational domain. 

% @Vu: to be written?
% joint decoding with a reconstruction model, e.g., Neural Machine Translation with Reconstruction (\cite{2016arXiv161101874T})

%%% Local Variables: 
%%% mode: latex
%%% TeX-master: "acl2017-relopt"
%%% End: 

\begin{table}
\centering
%\resizebox{\columnwidth}{!} {
\footnotesize
\begin{tabular}{|l||c|c|c|}
\hline
        & {\textbf{\# tokens}} & {\textbf{\# types}} & \textbf{\# sents}  \\
\hline \hline
\multicolumn{4}{|c|}{\textbf{BTEC zh$\rightarrow$en}} \\
\hline
%train & 421,536/454,448 & 3,436/3,121 &  44016  \\
%dev & 9,027/9,639 & 1,047/1,037 & 1006  \\
%test & 4,817/4,918 & ,712/,713 &  506  \\
train & 422k\ /\ 454k & 3k\ /\ 3k &  44,016  \\
dev & 10k\ /\ 10k & 1k\ /\ 1k & 1,006  \\
test & 5k\ /\ 5k & 1k\ /\ 1k &  506  \\
\hline
\multicolumn{4}{|c|}{\textbf{TED Talks de$\rightarrow$en}} \\
\hline
%train    & 4067k\ /\ 4329k & 26k\ /\ 19k &   194,181  \\
train    & 4m\ /\ 4m & 26k\ /\ 19k &   194,181  \\
dev-test2010     & 33k\ /\ 35k     & 4k\ /\ 3k &  1,565  \\
%test2013 & 22k\ /\ 23k     & 3k\ /\ 3k &  993  \\
test2014 & 26k\ /\ 27k     & 4k\ /\ 3k &  1,305  \\
%test2013\&14 & \ /\      & \ /\  &  2,298  \\
%train & 4067,432/4329,391 & 25,760/19,245 &   194181  \\
%dev & 33,417/35,134 & 3,874/3,285 &  1565  \\
%test2013 & 21,639/22,863 & 3,077/2,674 &  993  \\
%test2014 & 26,444/27,401 & 3,663/3,210 &  1305  \\
\hline
%\iffalse
%\multicolumn{8}{|c|}{\textbf{TED Talks fr$\rightarrow$en}} \\
%\hline
%train & 4926,511 & 4648,538 & 24,574 & 19,949 & 83,951 & 64,969 & 207323 & 1706 \\
%tune-tst2010 & 37,164 & 35,301 & 4,090 & 3,297 & ,613 & ,463 & 1664 & 11 \\
%test1-tst2013 & 25,343 & 23,765 & 3,377 & 2,807 & ,623 & ,515 & 1026 & 16 \\
%test2-tst2014 & 30,123 & 27,401 & 3,924 & 3,230 & ,687 & ,649 & 1305 & 15 \\
%\hline
%\fi
%\iffalse
\multicolumn{4}{|c|}{\textbf{WMT 2016 de$\rightarrow$en}} \\
\hline
%train & 99257,864 & 98759,816 & 90,338 & 78,172 & 0 & 0 & 4176062 & nil \\
train    & 107m\ /\ 108m & 90k\ /\ 78k &   4m  \\
dev-test2013\&14     & 154k\ /\ 152k     & 20k\ /\ 13k  &  6003  \\
test2015   & 54k\ /\ 54k    & 10k\ /\ 8k &   2169 \\
%tune-newstest2013 & & & & & & & & nil \\
%test-newstest2015 & & & & & & & & nil \\
%test-newstest2016 & & & & & & & & nil \\
\hline
%\multicolumn{9}{|c|}{\textbf{ACL WMT 2016 fr$\rightarrow$en}} \\
%\hline
%train & 106056,244 & 117762,470 & 89,066 & 82,138 &  & & 4622459 & nil \\
%tune-newstest2013 & & & & & & & & nil \\
%test-newstest2015 & & & & & & & & nil \\
%\hline
%\fi
\end{tabular}
%}
\caption{Statistics of the training and evaluation sets; token and types are 
presented for both source/target languages.  
%used,showing in each cell the count for the source language (left) and target language (right). ``\#types'' refers to the vocabulary counted on the training datasets;
%filtered vocabulary with word frequency cut-offs 5 for BTEC and TED Talks datasets, and converted vocabulary with BPE method for ACL WMT 2016 dataset; 
%``\#OOVs'' is count of out-of-vocabulary words of a given data set; ``nil'' means ``unavailable information''.
}
\label{tab:datasets}%-eval}
\end{table}

\section{Experiments}
\label{sec::exp}

\subsection{Setup}
%We conducted our experiments on various datasets with different scales, including: small (BTEC as in \cite{cohn-EtAl:2016:N16-1} Chinese $\leftrightarrow$ English), medium (IWSLT 2015 TED Talks \cite{cettoloEtAl:EAMT2012,Cettolo-iwslt-eval-camp-2014} German $\rightarrow$ English and French $\rightarrow$ English), and large (ACL WMT 2016\footnote{\url{http://www.statmt.org/wmt16/translation-task.html}} German $\rightarrow$ English and French $\rightarrow$ English). 
\paragraph{Datasets.} We conducted our experiments on datasets with different scales, translating between Chinese$\rightarrow$English using the BTEC corpus, and German$\rightarrow$English using the IWSLT 2015 TED Talks corpus \cite{Cettolo-iwslt-eval-camp-2014} and WMT 2016\footnote{\url{http://www.statmt.org/wmt16/translation-task.html}} corpus. 
%\cite{cettoloEtAl:EAMT2012,Cettolo-iwslt-eval-camp-2014}.
%
The statistics of the datasets can be found in Table~\ref{tab:datasets}.

\paragraph{NMT Models.}
We implemented our  continuous-optimisation based decoding method on top of the Mantidae toolkit\footnote{\url{https://github.com/duyvuleo/Mantidae}} \cite{cohn-EtAl:2016:N16-1}, and using the \emph{dynet} deep learning library\footnote{\url{https://github.com/clab/dynet}} \cite{2017arXiv170103980N}.
All neural network models were configured with 512 input embedding and hidden layer dimensions, and 256 alignment dimension, with 1 and 2 hidden layers in the source and target, respectively.
We used a LSTM recurrent structure \cite{Hochreiter:1997:LSM:1246443.1246450} for both source and target RNN sequences. 
For vocabulary sizes, we have chosen the word frequency cut-off 5 for creating the vocabularies for all datasets.
For large-scale dataset with WMT, we applied byte-pair encoding (BPE) method \cite{sennrich-haddow-birch:2016:P16-12} so that the neural MT system can tackle the unknown word problem \cite{luong-EtAl:2015:ACL-IJCNLP}.\footnote{With this BPE method,  the OOV rates of tune and test sets are lower than 1\%.} 
For training our neural models, the best perplexity scores on the development set is used for early stopping, which usually occurs after 5-8 epochs. 
%hyper-parameter(s) in extended NMT models

\begin{figure*}[t]
\begin{center}
  \makebox[\textwidth]{\includegraphics[width=0.75\paperwidth]{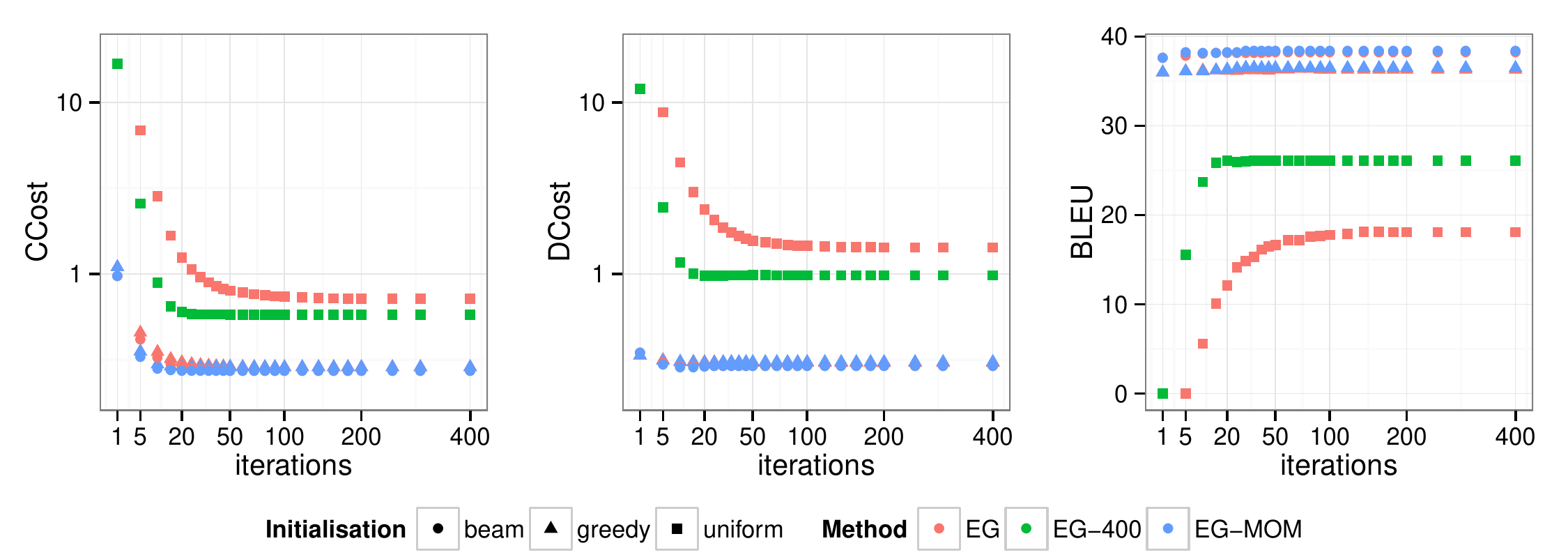}}
\end{center}
%\centering
%\includegraphics[]{plot-init.pdf} 
\caption{
    Analysis on effects of initialisation states (uniform vs. greedy vs. beam), step size annealing, momentum mechanism from BTEC zh$\rightarrow$en translation. \textbf{EG-400}: EG algorithm with step size $\eta=400$ (otherwise $\eta=40$); \textbf{EG-MOM}: EG algorithm with momentum.
} 
\label{fig:init}
\end{figure*}

\paragraph{Evaluation Metrics.} We evaluated in terms of search error, measured using the model score of the inferred solution (either continuous or discrete), as well as measuring the end translation quality with case-insensitive BLEU \cite{Papineni:2002:BMA:1073083.1073135}.  
The continuous cost measures $-\frac{1}{|\hat{\vy}|}\log P_{\vTheta} \left( \hat{\vy} \given \vx \right)$ under the model $\vTheta$; the discrete model score has the same formulation, albeit using the discrete rounded solution $\vy$ (see \S\ref{sec::prob_for}).
Note the cost can be used as a tool for selecting the best inference solution, as well as assessing convergence, as we illustrate below. 

\subsection{Results and Analysis}
\paragraph{Initialisation and Step Size.} 
As our relaxed optimisation problems are not convex, local optima are likely to be a problem.
We test this empirically, focusing on the effect that initialisation and step size, $\eta$, have on the inference quality. 

For plausible initialisation states, we evaluate different strategies: \emph{uniform} in which the relaxed variables $\hat{\vy}$ are initialised to $\frac{1}{|V_T|}$; and \emph{greedy} or \emph{beam} whereby  $\hat{\vy}$ are initialised based on an already good solution produced by a baseline decoder with greedy (gdec) or beam (bdec).
Instead of using the Viterbi outputs as a one-hot representation, we initialise to the probability prediction\footnote{Here, the EG algorithm uses $\operatorname{softmax}$ normalization whereas the SGD algorithm uses pre-$\operatorname{softmax}$ one.} vectors, which serves to limit attraction of the initialisation condition, which is likely to be a local (but not global) optima.

%uniform vs. greedy vs. beam search --> model difficulity of non-convexity and non-linearity of the objective function
%Figure~\ref{fig:init}

Figure~\ref{fig:init} illustrates the effect of initialisation on the EG algorithm, in terms of search error (left and middle) and translation quality (right), as we vary the number of iterations of inference.
There is clear evidence of non-convexity: all initialisation methods can be seen to converge using all three measures, however they arrive at highly different solutions.
Uniform initialisation is clearly not a viable approach, while greedy  and beam initialisation both yield much better results.
The best initialisation, beam, outperforms both greedy and beam decoding in terms of BLEU.  

%low convergence
Note that the EG algorithm has fairly slow convergence, requiring at least 100 iterations, irrespective of the initialisation.
%momentum
To overcome this, we use momentum \cite{Qian1999145} to accelerate the convergence by modifying the term $\nabla_{i,w}^{t}$ in Algorithm~\ref{alg1a} with a weighted moving average of past gradients:
\begin{align*}
\nabla_{i,w}^{t-1} = \gamma \nabla_{i,w}^{t-2} + \eta \frac{\partial Q(\hat{\vy}_1^{t-1},\ldots,\hat{\vy}_{\ell}^{t-1})}{\partial \hat{\vy}_i(w)} 
\end{align*} 
where we set the momentum term $\gamma=0.9$.
The EG with momentum (\textbf{EG-MOM}) converges after fewer iterations (about 35), and results in marginally better BLEU scores. The momentum technique is usually used for SGD involving additive updates; it is  interesting to see  it also works in EG with multiplicative updates.

%step size
The step size, $\eta$, %\footnote{We call step size instead of learning rate for optimisation in decoding/inference.} 
is another important hyper-parameter for gradient based search.
%
%there is in fact no obvious method to choose an appropriate step size for gradient descent - based optimisation; instead, depending on a specific modeling task. 
%
We tune the step size using line search over $[10, 400]$ over the development set.
%\footnote{The step size here is much bigger than usual used in training, possibly because in terms of the optimisation for inference, we may need a large learning rate to compensate for small gradients.} 
%
Figure~\ref{fig:init} illustrates the effect of changing step size from 50 to 400 (compare \textbf{EG} and \textbf{EG-400} with \textbf{uniform}), which results in a marked difference of about 10 BLEU points, underlining the importance of tuning this value.
%
%It could be due to the fact that uniform initialisation causes very small values of gradients, and large step sizes are likely to compensate for it. 
%
We found that EG with momentum had less of a reliance on step size, with optimal values in $[10,50]$; we use this setting hereafter.
%@Vu: perhaps requires add more lines in plot-init for the effects of step sizes?

\begin{figure}
\centering
\includegraphics[width=0.9\columnwidth]{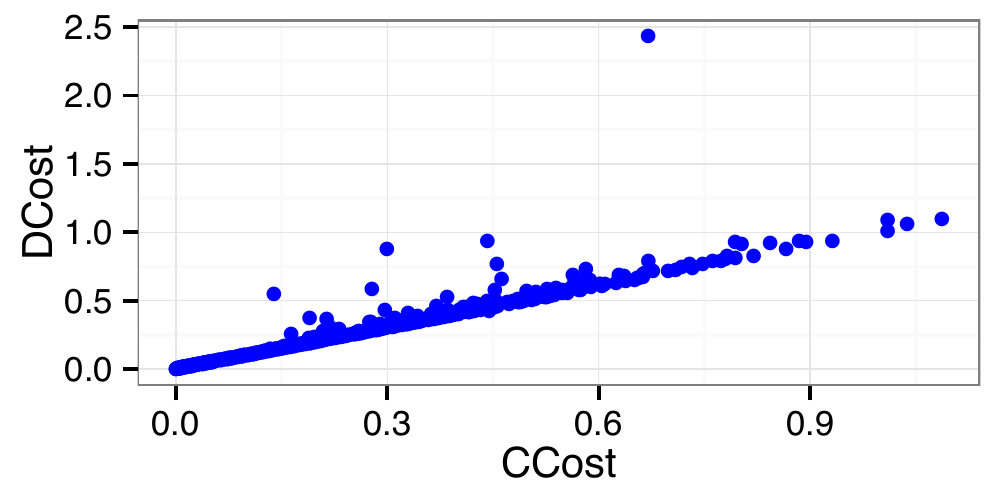} 
\caption{
    Comparing discrete vs continuous costs from BTEC zh$\rightarrow$en translation, using the EG algorithm with momentum, $\eta=50$. Each point corresponds to a sentence. 
} 
\label{fig:ccost_vs_dcost}
\end{figure}

\paragraph{Continuous vs Discrete Costs.} 
Another important question is whether the assumption behind continuous relaxation is valid, i.e., if we optimise a continuous cost to solve a discrete problem, do we improve the discrete output?
Although the continuous cost diminishes with inference iterations (Figure~\ref{fig:init} centre), and appears to converge to an optima, it is not clear whether this corresponds to a better discrete output (note that the BLEU scores do show improvements Figure~\ref{fig:init}.)
Figure~\ref{fig:ccost_vs_dcost} illustrates the relation between the two cost measures, showing that in almost all cases the discrete and continuous costs are identical.
Linear relaxation effectively fails only for a handful of cases, where the nearest discrete solution is significantly worse than it would appear using the continuous cost.   

%(optional) ccost vs dcost

\begin{figure}
    \centering
\includegraphics[width=\columnwidth]{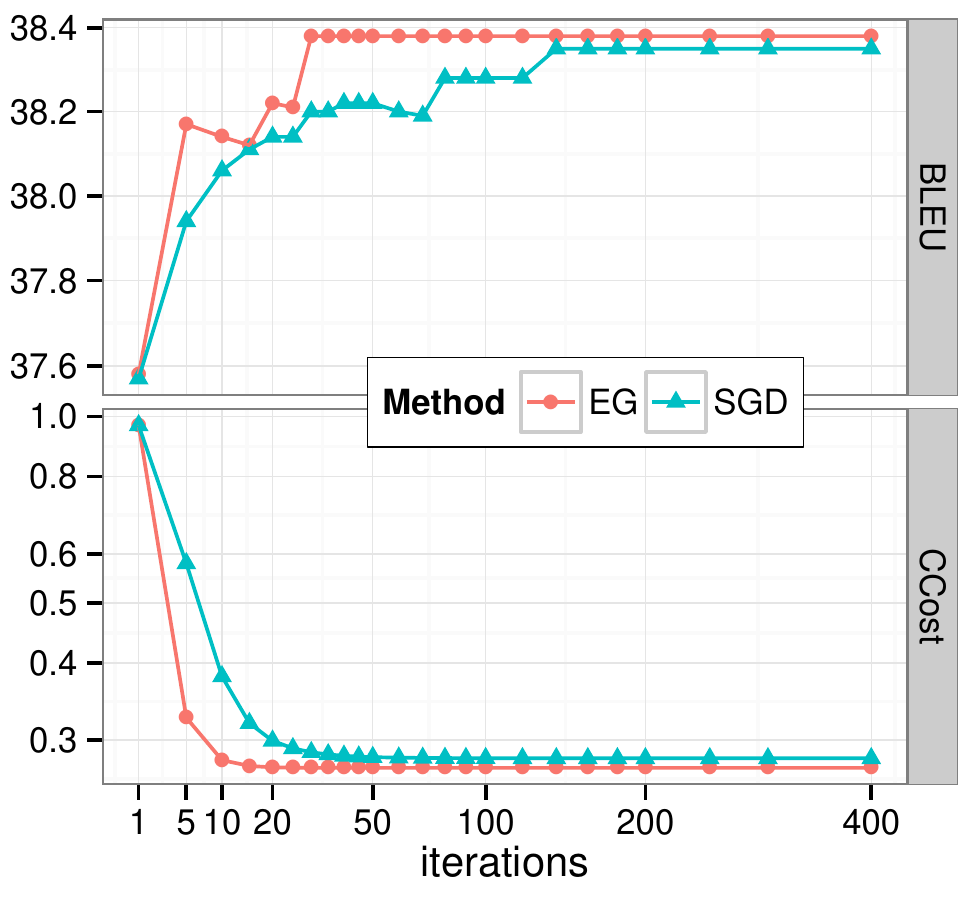} 
    \caption{
    Analysis on convergence and performance comparing SOFTMAX and EG algorithms from BTEC zh$\rightarrow$en translation. Both algorithms use momentum and step size 50.
} 
\label{fig:softmax_eg}
\end{figure}

\paragraph{EG vs SGD.} 
%There is no significant difference between the two algorithms EG and SGD, as well as no clear evidence that which algorithm works better. 
Both the EG and SGD algorithms are iterative methods for solving the relaxed optimisation problem with simplex constraints.
We measure empirically their difference in terms of quality of inference and speed of convergence, as illustrated in Figure~\ref{fig:softmax_eg}.
% the convergence speed and BLEU scores of the two algorithms. 
Observe that SGD requires 150 iterations for convergence, whereas EG requires many fewer (50). 
This concurs with previous work on learning structured prediction models with EG \cite{Globerson:2007:EGA:1273496.1273535}.
Further, the EG algorithm consistently produces better results in terms of both model cost and BLEU. 

% SOFTMAX vs. EG

\newcolumntype{R}[2]{%
    >{\adjustbox{angle=#1,lap=\width-(#2)}\bgroup}%
    l%
    <{\egroup}%
}
\newcommand*\rot{\multicolumn{1}{R{0}{1em}}}
\newcommand*\rotl{\multicolumn{1}{R{75}{1em}|}}
\newcommand*\rotll{\multicolumn{1}{|R{75}{1em}||}}

\newcommand{\greedydec}{gdec}
\newcommand{\beamdec}{bdec}
\newcommand{\optidec}{EGdec}
\newcommand{\furerank}{full rerank}
\newcommand{\firerank}{filtered rerank}

\newcommand{\zhen}{BTEC}
\newcommand{\tdeen}{TEDTalks}
\newcommand{\wdeen}{WMT}

\newcommand{\bleu}{BLEU}
\newcommand{\len}{AvgLen}

\newcommand{\bestss}{\ensuremath{^{\dagger}}}
%\newcommand{\ubold}{\fontseries{b}\selectfont} % renew def. for non-extended bold font
%\robustify\ubold

\newcommand{\boldentry}[1]{%
  \multicolumn{1}{S[table-format=2.2,
                    mode=text,
                    text-rm=\fontseries{b}\selectfont
                   ]}{#1}}

\begin{table}
{%\footnotesize
\sisetup{
round-mode = places,
round-precision = 2,
table-format = 2.2,
table-space-text-post = {\bestss},
}%
\begin{center}
\begin{tabular}{lSS}
\toprule
 & {\bleu} & {\len} \\ 
\toprule
%\midrule 
%$\textrm{\beamdec}_{\textrm{left-to-right}}$ & 26.69 & 20.73 \\%44,959 \\
%$\textrm{\firerank}_{\textrm{bidirectional}}$   & 26.84 &  20.66 \\ %44,804 \\
%$\textrm{\furerank}_{\textrm{bidirectional}}$  & 27.34  & 21.76 \\ %47,192 \\
%$\textrm{\optidec}_{\textrm{bidirectional}}$ 	& & \\
%\ \ \ \  w/ {\textrm{beam init}} & 27.34 & 20.73 \\ %44,965 \\
%\ \ \ \  w/ {\textrm{rerank init}} & 27.78{\bestss}  & 21.70 \\ %47,059 \\
$\textrm{\beamdec}_{\textrm{left-to-right}}$ & 26.69 & 20.73 \\%44,959 \\
\firerank   & 26.84 &  20.66 \\ %44,804 \\
\optidec \ w/ beam init & 27.34 & 20.73 \\ %44,965 \\
\bottomrule 
\furerank  & 27.34  & 21.76 \\ %47,192 \\
\optidec  \ w/ rerank init & 27.78{\bestss}  & 21.70 \\ %47,059 \\

%$\textrm{\optidec}_{\textrm{bidirectional}}$ 	& & \\
%\ \ \ \  w/ {\textrm{beam init}} & 27.34 & 20.73 \\ %44,965 \\
%\ \ \ \  w/ {\textrm{rerank init}} & 27.78{\bestss}  & 21.70 \\ %47,059 \\

\bottomrule 
\end{tabular}
\end{center}
}
\caption{The BLEU evaluation results with EG algorithm against 100-best reranking on WMT evaluation dataset; \bestss: best performance.}
\label{tab:result_eg_vs_reranking}
\end{table}

\paragraph{EG vs Reranking.} 
Reranking is an alternative method for integrating global factors into the existing NMT systems. 
We compare our EG decoding algorithm against the reranking approach with bidirectional factor where the N-best outputs of a left-to-right decoder is re-scored with the forced decoder operateing in a right-to-left fashion. The results are shown in Table~\ref{tab:result_eg_vs_reranking}. 
Our EG algorithm initialised with the reranked output achieves the best  BLEU score.  
We also compare reranking with EG algorithm initialised with the beam decoder, where for direct comparison we filter out sentences with length greater than that of the beam output in the k-best lists.   
%
%We find that our EG algorithm with beam initialisation is comparable to full reranking (27.34 vs 27.34) which exploits all sentences in the nbest list; although EG algorithm is limited only in search space for beam output.  
%For direct comparison, we limit the reranking in which we filter sentences with lengths greater than beam output, resulting in worse result (26.84 vs 27.34).
%Our EG algorithm with initialisation over reranked output further gives significantly better BLEU score (27.78). 
These results show  that the EG algorithm is capable of effectively exploiting the search space. 

As opposed to re-ranking, our approach does not need a pipeline, e.g. to produce and score $n$best lists, to tune the weights etc. 
% with MERT/MIRA, etc. 
% 
The run time complexity of our approach is comparable with that of reranking: our model needs repeated application of the  NMT global factors to navigate through the search space, whereas re-ranking needs to use the underlying NMT model to generate the the $n$best list and generate their global NMT scores. % based on the  while exploring the search space. 
%
%Additionally, re-ranking needs to generate the score of candidate translations in the $n$best list using additional NMT factors, both require a surrogate model (for initialisation or generating $n$best lists) and application of  NMT factors models to several different decoding sequences (i.e., k vs \#iterations of EG). 
Note that our method swaps some sparse 1-hot vector operations for dense and requires a back-propagation pass, where both operations are relatively cheap on GPUs. 
Overall our expectation is that for sufficiently large k to get improvements in BLEU, our relaxed decoding is potentially faster.

\paragraph{Computational Efficiency.} 
We also quantify the computational efficiency of the proposed decoding approach. 
Benchmarking on a GPU Titan X for decoding 506 sentences of BTEC zh$\rightarrow$en, it takes 0.02s/sentence for greedy,  0.07s/sentence for beam 5, 0.11s/sentence for beam 10, and 3.1s/sentence for our relaxed EG decoding (with an average of 35 EG iterations). 
More concretely, our relaxed EG decoding includes: 0.94s/sentence for the forward step, 2.09s/sentence for the backward step, and $<$0.01s/sentence for the update and additional steps. 
It turns out that the backward step is the most computationally expensive step, limiting the practical applicability of the proposed decoding approach. 
Addressing this important issue is left for our future research. 

%%%%%%%%%%%%%%%%%%%%%%%%%%%%%%%%%%%%%%%%%%%%%%%%%%%

\begin{table}
{%\footnotesize
\sisetup{
round-mode = places,
round-precision = 2,
table-format = 2.2,
table-space-text-post = {\bestss},
}%
\begin{center}
%\begin{tabular}{lSSSSSS}
\begin{tabular}{lSSS}
\toprule
 %& {\zhen} & {\enzh}& {\tdeen}& {\tende}& {\fdeen}& {\fende} \\ 
 & {\zhen} & {\tdeen} & {\wdeen} \\ 
 & ${_{\textrm{zh} \rightarrow \textrm{en}}}$ &  ${_{\textrm{de} \rightarrow \textrm{en}}}$ &  ${_{\textrm{de} \rightarrow \textrm{en}}}$ \\ 
 
%&\rot{\zhen} & \rot{\enzh}& \rot{\tdeen}& \rot{\tende}& \rot{\fdeen}& \rot{\fende} \\ 
\toprule
%$\textrm{\greedydec}_{\textrm{left-to-right}}$  & 35.98 & 20.23 & 27.58 & 23.16 & 23.16 & 19.33 \\
$\textrm{\greedydec}_{\textrm{left-to-right}}$  & 35.98 & 23.16 & 24.41 \\
%$\textrm{\greedydec}_{\textrm{right-to-left}}$  & 35.86 & 20.01 & 26.83 & 20.78 & 21.95 & 17.18 \\
$\textrm{\greedydec}_{\textrm{right-to-left}}$  & 35.86 & 21.95 & 23.59 \\
%$\textrm{\optidec}_{\textrm{ginit}}$      		& 36.34 & 20.42 & 27.81 & 23.29 & 23.28 & 19.37 \\
$\textrm{\optidec}_{\textrm{greedy init}}$      		& 36.34 & 23.28 & 24.63 \\
%\ \ \ \ +bidirectional                          & \boldentry{36.67} & \boldentry{21.50\textnormal{\bestss}} & \boldentry{28.65\textnormal{\bestss}} & %\boldentry{23.76} & \boldentry{23.91} & \boldentry{19.92} \\
\ \ \ \ +bidirectional                       & \boldentry{36.67} & \boldentry{23.91} &  \boldentry{25.37\textnormal{\bestss}} \\
%\ \ \ \ +bilingual                              & \boldentry{36.88\textnormal{\bestss}} & \boldentry{21.16} & \boldentry{28.41} & %\boldentry{24.07\textnormal{\bestss}} & \boldentry{24.01\textnormal{\bestss}} & \boldentry{20.22\textnormal{\bestss}} \\
\ \ \ \ +bilingual                      & \boldentry{36.88\textnormal{\bestss}} & \boldentry{24.01\textnormal{\bestss}} & \boldentry{25.21}  \\
%\ \ \ \ +bilingual+trace                        & \boldentry{36.59} & \boldentry{21.50\textnormal{\bestss}} & \boldentry{28.23} & \boldentry{23.82} & \boldentry{23.65} &  \boldentry{19.85} \\
\midrule 
%$\textrm{\beamdec}_{\textrm{left-to-right}}$    & 38.02 & 21.20 & 28.74 & 24.62 & 23.95 & 20.64 \\
$\textrm{\beamdec}_{\textrm{left-to-right}}$    & 38.02 & 23.95 & 26.69  \\
%$\textrm{\beamdec}_{\textrm{right-to-left}}$    & 37.38 & 17.60 & 28.30 & 22.17 & 23.13 & 19.45 \\
$\textrm{\beamdec}_{\textrm{right-to-left}}$    & 37.38  & 23.13  & 26.11  \\
%$\textrm{\optidec}_{\textrm{binit}}$     		& 38.38 & 21.11 & 28.75 & 24.58 & 24.02 & 20.54 \\
$\textrm{\optidec}_{\textrm{beam init}}$     		& 38.38 & 24.02 &  26.66 \\
%\ \ \ \ +bidirectional                          & \boldentry{39.13\textnormal{\bestss}} & \boldentry{21.87} & \boldentry{29.92\textnormal{\bestss}} & \boldentry{25.18} & \boldentry{24.72\textnormal{\bestss}} &  21.06\textnormal{\bestss} \\
\ \ \ \ +bidirectional                        & \boldentry{39.13\textnormal{\bestss}} & \boldentry{24.72\textnormal{\bestss}}  & \boldentry{27.34\textnormal{\bestss}} \\
%\ \ \ \ +bilingual     						    & 38.25 & \boldentry{21.86} & \boldentry{29.20} & \boldentry{25.19\textnormal{\bestss}} & \boldentry{24.60} & 21.01 \\
\ \ \ \ +bilingual					    & 38.25 & \boldentry{24.60}  & 26.82 \\
%\ \ \ \ +bilingual+trace					    & \boldentry{38.64} & \boldentry{22.10\textnormal{\bestss}} & 28.87 & \boldentry{25.09} & \boldentry{24.52} & 20.57 \\
\bottomrule 
\end{tabular}
\end{center}
}
\caption{The BLEU evaluation results across evaluation datasets for EG algorithm variants against the baselines; \textbf{bold}: statistically significantly better than the best greedy or beam baseline, \bestss: best performance on dataset.}
\label{tab:result_all}
\end{table}

%%% Local Variables: 
%%% mode: latex
%%% TeX-master: "acl2017-relopt"
%%% End: 

\paragraph{Main Results.} 
% all results
% relopt vs. vanilla
%explain \ref{tab:result_all}, with greedy init we see small improvements in bleu for all cases (except en->de2014 and en->de 2013). 
Table~\ref{tab:result_all} shows our experimental results across all datasets, evaluating the EG algorithm and its variants.\footnote{Given the aforementioned analysis and space constraints, here we reported the results for the EG algorithm only.}
For the EG algorithm with greedy initialisation (top), we see small but consistent improvements in terms of BLEU. 
Beam initialisation led to overall higher BLEU scores, and again demonstrating a similar pattern of improvements, albeit of a lower magnitude, over the initialisation values.
%

% relopt with extended NMT models
%with our method, we can decode in extended nmt models (i.e. bilingual ensemble or bidirectional ensemble)  and we get statistically significant BLEU improvement (VU: please look at the statistical significance of BLEU score)
Next we evaluate the capability of our inference method with extended NMT models, where approximate algorithms such as greedy or beam search are infeasible.
With the \textbf{bidirectional} ensemble, we obtained the statistically significant BLEU score improvements compared to the unidirectional models, for either greedy or beam initialisation. 
This is interesting in the sense that the unidirectional right-to-left model always performs worse than the left-to-right model. % (sometimes badly, e.g., en$\rightarrow$de'13). 
However, our method with bidirectional ensemble is capable of combining their strengths in a unified setting. % rather than via re-ranking evaluation. 
For the \textbf{bilingual} ensemble, we see similar effects, with better BLEU score improvements in most cases,  albeit of a lower magnitude, over the bidirectional one. 
%
%Further, we evaluated the effect of adding trace bonus for the \textbf{bilingual} ensemble. 
%
%Extending inference to use a trace bonus is not always beneficial, leading to a small degradation in performance in several cases. 
%
This is likely to be due to a disparity with the training condition for the models, which were learned independently of one another. %, without the trace bonus term.
%It is likely that there exist both agreement as well as dis-agreement between the two models %(source-to-target and target-to-source) 
%alignments which causes ambiguity for the model. 

Overall, decoding in extended NMT models leads to performance improvements compared to the baselines. 
This is one of the main findings in this work, and augurs well for its extension to other global model variants.

\begin{figure*}
\centering
\includegraphics[width=2.0\columnwidth]{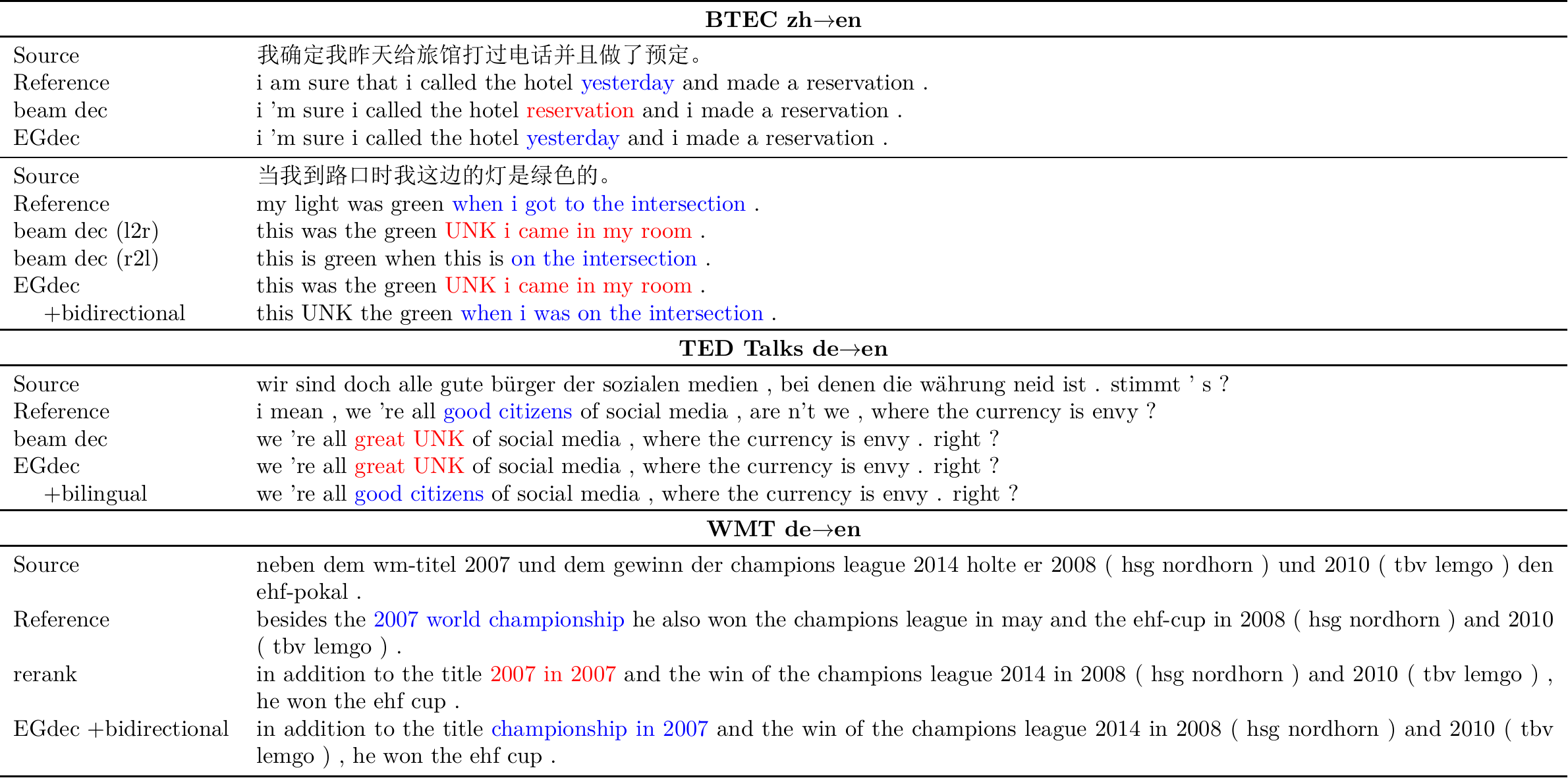} 
\caption{Translation examples generated by the models.} 
\label{fig:trans_exp}
\end{figure*}

%@Vu: need to find a space to show translation example(s) (maybe a table?)
%...

%%% Local Variables: 
%%% mode: latex
%%% TeX-master: "acl2017-relopt"
%%% End: 

\section{Related Work}
\label{sec::rel}

Decoding (inference) for neural models is an important task; however, there is limited research in this space perhaps due to the challenging nature of this task, with only a few works exploring some extensions to improve upon them.
The most widely-used inference methods include sampling \cite{2016arXiv160503835C}, greedy and beam search \cite[\emph{inter alia}]{Sutskever:2014:SSL:2969033.2969173,bahdanau:ICLR:2015}, and reranking \cite{sennrich-edinburgh-wmt16,2016arXiv160100372L}.

%NPAD
\newcite{2016arXiv160503835C} proposed to perturb the neural model by injecting noise(s) in the hidden transition function of the conditional recurrent neural language model during greedy or beam search, and execute multiple parallel decoding runs. 
This strategy can improves over greedy and beam search; however, it is not clear how, when and where noise should be injected to be beneficial. 
%
%beam search during training 
Recently, \newcite{wiseman-rush:2016:EMNLP2016} proposed  beam search optimisation while \emph{training}  neural models, where 
the model parameters are updated in case the gold standard falls outside of the beam.
This exposes the model to its past incorrect predicted labels, hence making the training more robust. %\footnote{See also imitation learning, which supports training-time conditioning on model errors to allow more robust test inferences with greedy search \cite[\emph{inter alia}]{chang2015learning,ballesteros-EtAl:2016:EMNLP2016}.}
This is orthogonal to our approach where  we focus on the decoding problem with a pre-trained model.
%Further, our continuous-optimisation based  decoding  is more direct in accounting for interdependencies among the predicted variables compared to the beam search training which tries to account for these dependencies indirectly through the future cost component. 

%reranking \cite{sennrich-edinburgh-wmt16,2016arXiv160100372L,2016arXiv161108562L}
Reranking has also been proposed as a means of global model combination: \newcite{sennrich-edinburgh-wmt16} and  \newcite{2016arXiv160100372L}  re-rank the left-to-right decoded translations based on the scores of a right-to-left model, learning to more diverse translations.
Related, \newcite{2016arXiv161108562L} learn to adjust the beam diversity with reinforcement learning.

%master thesis 
Perhaps most relevant is \newcite{Snelleman-master-thesis}, performed concurrently to this work, who also proposed an inference method for NMT using linear relaxation.
\citeauthor{Snelleman-master-thesis}'s method was similar to our SGD approach, however he did not manage to outperform beam search baselines with an encoder-decoder.
In contrast we go much further, proposing the EG algorithm, which we show works much more effectively than SGD, and demonstrate how this can be applied to inference in an attentional encoder-decoder.
Moreover, we demonstrate the utility of related optimisation for inference over global ensembles of models, resulting in consistent improvements in search error and end translation quality.
%

% Concurrent to us, \newcite{Snelleman-master-thesis} has presented a preliminary work on neural translation decoding with gradient descent; however, our work is different in several aspects. 
% %
% Firstly, we have proposed two different optimisation algorithms including EG, which are more general than  \newcite{Snelleman-master-thesis}'s proposed algorithms. 
% %
% Secondly, we have shown that the continuous optimisation based decoding is applicable to extended NMT, leading to improved results.
% %  
% Lastly, our experiments have been conducted with an attentional NMT, instead of pure encoder-decoder models as in \newcite{Snelleman-master-thesis}. 

% relaxed technique
Recently, relaxation techniques have been applied to deep  models for training and inference in text classification \cite{Belanger:2016:SPE:3045390.3045495,2017arXiv170305667B}, and  fully differentiable training of sequence-to-sequence models
with scheduled-sampling \cite{Goyalacl17}.
Our work has applied the relaxation technique specifically for \emph{decoding} in NMT models. 

%%% Local Variables: 
%%% mode: latex
%%% TeX-master: "acl2017-relopt"
%%% End: 

\section{Conclusions}
\label{sec::con}

This work presents the first attempt in formulating  decoding in NMT as a continuous optimisation problem.
%
%of a novel framework for decoding in neural translation models, where  decoding  
%is formulated as  a continuous optimisation problem. 
%
The core idea is to drop the integrality (i.e. one-hot vector) constraint  from the prediction variables and allow them to have 
soft assignments within the probability simplex while minimising the loss function produced by the neural model.
%treat them by 
%soft assignments belonging to the probability simplex with the goal of minimising the loss produced by the neural model. 
%
%The goal of the decoding is then to minimise the loss of the neural model for these soft assignments.
%
We have provided two optimisation algorithms -- exponentiated gradient (EG) and stochastic gradient descent (SGD) -- 
for optimising the resulting contained optimisation problem, where our findings show the effectiveness of EG compared to SGD. 
%
%Since our approach is not dependent on any decoding direction, it provides a flexible and powerful 
%framework to decode in neural models involving outputs with complex interdependencies, such as those in machine translation. 
%
Thanks to our framework, we have been able to decode when intersecting left-to-right and right-to-left  as well as source-to-target and target-to-source NMT models. %, where the popular greedy/beam search algorithms are not applicable.
Our results show that our decoding framework is effective and lead to substantial improvements in translations\footnote{Some comparative translation examples are included in Figure~\ref{fig:trans_exp}.} generated from 
the intersected models, where the typical greedy or beam search algorithms are not applicable. 

%future works
This work raises several compelling possibilities which we intend to address in future work, including improving the decoding speed,
integrating additional constraints such as word coverage and fertility into \emph{decoding},\footnote{These constraints have
only been used for training in the previous works \cite{cohn-EtAl:2016:N16-1,mi-EtAl:2016:EMNLP2016}.} and applying our method to other intractable structured prediction such as parsing.

% First, do these improvements carry to larger-scale datasets (e.g., ACL WMT)?
%1)  improving the method with faster run-time of inference, in order to make the method practical for widespread use; 
% through parallelisation or via different optimisation methods?
%, such as the bundle method \cite{2016arXiv160907152A}.
%2) integerating additional constraints, such as word coverage, fertility etc into \emph{decoding}, which previously have only been included during training \cite{cohn-EtAl:2016:N16-1,mi-EtAl:2016:EMNLP2016}; 
%3) and applying the method to other intractable structured prediction tasks beyond translation with more strict constraints, such as: parsing, summarisation.%, sequence tagging and network labelling, and for inference with classical non-neural architectures.

%%% Local Variables: 
%%% mode: latex
%%% TeX-master: "acl2017-relopt"
%%% End: 

\section*{Acknowledgments} 

We thank the reviewers for valuable feedbacks and discussions. 
Cong Duy Vu Hoang is supported by Australian Government Research Training Program Scholarships at the University of Melbourne, Australia. Trevor Cohn is supported by the ARC Future Fellowship. 
This work is partly supported by an ARC DP grant to Trevor Cohn and Gholamreza Haffari. 

\bibliography{emnlp2017-relopt}
\bibliographystyle{emnlp_natbib}

\end{document}